\documentclass[11pt]{article}

\usepackage[margin=1in]{geometry}
\PassOptionsToPackage{table,dvipsnames}{xcolor}
\usepackage{graphicx}
\usepackage[dvipsnames]{xcolor}
\usepackage{comment}
\usepackage{multirow}
\usepackage{array}
\usepackage{adjustbox}
\usepackage{siunitx}
\usepackage{colortbl}
\usepackage{booktabs}
\usepackage{caption}
\definecolor{green1}{RGB}{198, 239, 206}
\definecolor{green2}{RGB}{218, 239, 225}
\definecolor{yellow1}{RGB}{255, 235, 156}
\definecolor{yellow2}{RGB}{255, 242, 204}
\definecolor{orange1}{RGB}{255, 199, 206}
\definecolor{orange2}{RGB}{255, 229, 204}
\definecolor{headergray}{gray}{0.9}
\usepackage{pgfplots}
\pgfplotsset{compat=1.18}
\usepackage{tikz}

\usepackage[utf8]{inputenc}
\usepackage[english]{babel}
\usepackage[T1]{fontenc}
\usepackage{lmodern}
\usepackage{csquotes}

\usepackage[
  backend=biber,
  sorting=nyt,
  style=authoryear,
  sortcites=true,
  maxcitenames=2,
  mincitenames=1,
  maxbibnames=10,
  useprefix=true,
  date=year,
  url=false,
  urldate=long,
  isbn=false,
  giveninits=true,
  doi=false,
  terseinits=false,
  autocite=inline,
  uniquename=false,
  dashed=false,
  uniquelist=false,
]{biblatex}
\IfFileExists{arxiv.bbl-only}{}{%
  \IfFileExists{references.bib}{\addbibresource{references.bib}}{}%
}

\title{Sustainability via LLM Right-sizing}
\date{April 17, 2025}

\definecolor{darkblue}{RGB}{0,0,128}
\usepackage{hyperref}
\hypersetup{colorlinks=true, linkcolor=darkblue, citecolor=darkblue, urlcolor=darkblue}

\usepackage{tabularx}
\usepackage{pifont}

\newcommand{\greencheck}{\textcolor{green}{\ding{51}}}
\newcommand{\redcross}{\textcolor{red}{\ding{55}}}

\newcommand{\redTodo}[1]{\textcolor{red}{[#1]}}

\usepackage{listings}

\definecolor{quotebg}{RGB}{245,245,245}     
\definecolor{quoteborder}{RGB}{160,160,160} 

\lstdefinestyle{feedbackprompt}{
  basicstyle=\footnotesize\ttfamily,
  backgroundcolor=\color{quotebg},
  frame=single,
  framesep=8pt,
  framerule=1pt,
  rulecolor=\color{quoteborder},
  breaklines=true,
  breakatwhitespace=false,
  captionpos=b,
  showstringspaces=false,
  numbers=none,
  xleftmargin=5pt,
  xrightmargin=5pt,
  aboveskip=10pt,
  belowskip=10pt
}

\begin{document}

\begin{center}
  {\LARGE\bfseries Sustainability via LLM Right-sizing\par}
  \vspace{0.35em}
  {\large\textit{April 17, 2025}\par}
  \vspace{1.25em}
\end{center}

\begin{center}
\begin{tabularx}{\textwidth}{@{}XX@{}}
\begin{center}
\Large
\textbf{Jennifer Haase}\\
\normalsize
Weizenbaum Institute and Humboldt-University zu Berlin\\
Berlin, Germany\\
\url{jennifer.haase@hu-berlin.de}
\end{center}
&
\begin{center}
\Large
\textbf{Finn Klessascheck}\\
\normalsize
TU Munich and Weizenbaum Institute, Germany\\
Heilbronn and Berlin, Germany\\
\url{finn.klessascheck@tum.de}
\end{center}
\\[0.5cm]
\begin{center}
\Large
\textbf{Jan Mendling}\\
\normalsize
Weizenbaum Institute and Humboldt-University zu Berlin\\
Berlin, Germany\\
\url{jan.mendling@hu-berlin.de}
\end{center}
&
\begin{center}
\Large
\textbf{Sebastian Pokutta}\\
\normalsize
TU Berlin and Zuse Institute Berlin\\
Berlin, Germany\\
\url{pokutta@zib.de}
\end{center}
\end{tabularx}
\end{center}

\begin{abstract}
 Large language models (LLMs) have become increasingly embedded in organizational workflows. This has raised concerns over their energy consumption, financial costs, and data sovereignty. While performance benchmarks often celebrate cutting-edge models, real-world deployment decisions require a broader perspective: when is a smaller, locally deployable model ``good enough''? This study offers an empirical answer by evaluating eleven proprietary and open-weight LLMs across ten everyday occupational tasks, including summarizing texts, generating schedules, and drafting emails and proposals. Using a dual-LLM-based evaluation framework, we automated task execution and standardized evaluation across ten criteria related to output quality, factual accuracy, and ethical responsibility. Results show that GPT-4o delivers consistently superior performance but at a significantly higher cost and environmental footprint. Notably, smaller models like Gemma-3 and Phi-4 achieved strong and reliable results on most tasks, suggesting their viability in contexts requiring cost-efficiency, local deployment, or privacy. A cluster analysis revealed three model groups --premium all-rounders, competent generalists, and limited but safe performers-- highlighting trade-offs between quality, control, and sustainability. Significantly, task type influenced model effectiveness: conceptual tasks challenged most models, while aggregation and transformation tasks yielded better performances. We argue for a shift from performance-maximizing benchmarks to task- and context-aware sufficiency assessments that better reflect organizational priorities. Our approach contributes a scalable method to evaluate AI models through a sustainability lens and offers actionable guidance for responsible LLM deployment in practice.

  \emph{\textbf{Keywords:} Large Language Models, benchmarking, task performance, sustainability}
\end{abstract}

\section{Introduction}

Artificial Intelligence (AI) is becoming an integral component of business and organizational processes \parencite{raischCombiningHumanArtificial2024}, transforming how decisions are made \parencite{delvalleAIpoweredRecommenderSystems2024}, services are delivered \parencite{huiWhenServiceQuality2024}, and tasks are automated \parencite{haaseInterdisciplinaryDirectionsResearching2024}. This rapid integration has sparked growing concerns over energy consumption, escalating operational costs, and dependence on external cloud providers \parencite{fioravanteBusinessCaseResponsible2024, hoganAIHotMess2024}. As organizations face increasing regulatory pressure and environmental accountability, the question of how to deploy AI systems sustainably is gaining urgency \parencite{tripathiEthicalPracticesArtificial2025,verdecchiaASystematicReview2023}.

Sustainability in the context of AI not only concerns carbon emissions but also issues of data sovereignty and operational costs. First, the energy demand of large-scale AI models contributes significantly to CO\textsubscript{2} emissions, with model size and inference compute being key drivers. Second, many AI solutions rely on centralized cloud infrastructure, raising questions about data privacy, ownership, and data residency. Third, the financial cost of running large models can be prohibitive, especially when usage is scaled across organizational tasks. Here, local models promise significantly reduced cost-per-token, often down to a fraction of a cent.

Despite increasing awareness of these challenges, existing AI evaluation measures remain narrowly focused on technical performance metrics such as accuracy, scalability, and generalization \parencite{liCrowdsourcedDataHighQuality2024, whiteLiveBenchChallengingContaminationFree2024}. These benchmarks, while important, rarely consider the full spectrum of real-world deployment constraints. Crucially, professionals lack guidance on a core question: When is a smaller, locally deployed model ``good enough'' to perform a task effectively, affordably, and autonomously? And how should different AI models be selected for various types of tasks, considering trade-offs in sustainability, sovereignty, and cost?

To address this research problem, we tested 11 LLMs across 10 typical office tasks (like writing an email, summarizing a text, conceptualizing an idea) and evaluated the individual outputs through LLMs. This approach enables the empirical comparison of LLMs, ranging from powerful, cloud-hosted systems to lightweight, local alternatives, across a set of representative occupational tasks. We ask two central research questions: (1) Can smaller, locally deployed models effectively perform occupational tasks? and (2) Which models are most suitable for which types of tasks? Our analysis incorporates core quality aspects of the output generated, its factual integrity, and ethical and social responsibility. We further discuss sustainability metrics---including compute-based CO\textsubscript{2} proxies, deployment location (local vs.~cloud), and cost per million tokens---to evaluate AI readiness in practical organizational contexts. In doing so, we contribute to the growing discourse in \emph{Green Information Systems} (Green IS) and propose a structured approach to more sustainable LLM adoption.

The remainder of this paper first discusses the sustainability aspects from the Green IS perspective, provides a brief analysis of the technical capabilities of different LLMs, describes our LLM-based evaluation method, and presents empirical results on model performance across varied deployment scenarios. We conclude with a discussion of practical implications, limitations, and future directions for sustainable AI integration in organizations.

\section{Background}

To situate our contribution, this section outlines key debates on sustainable AI deployment in organizations. We build on the triad model of sustainability, discuss technical and economic considerations in LLM tooling, and argue for a shift toward task- and deployment-aware benchmarking practices. Taken together, these strands motivate the need for evaluating AI systems not only in terms of raw capability, but also in terms of operational feasibility and responsible adoption.

\subsection{Sustainability and AI Deployment}
\label{sec:lit}

The Information Systems (IS) community increasingly recognizes sustainability as a critical imperative, particularly through the subfield of Green IS, which seeks to leverage information technology for environmental and social benefit \autocite{seidelSustainabilityImperativeInformation2017, watsonInformationSystemsEnvironmentally2010, kirchner-krathChallengesAdoptionSustainability2024}. Green IS emphasizes practices and processes enabled by digital systems that improve sustainability within organizational and societal contexts \autocite{veitDigitalizationProblemSolution2023}. Within this discourse, two contrasting roles of digital technologies are discussed: On the one hand, technology can exacerbate environmental degradation and social inequality through resource-intensive systems. On the other hand, it is positioned as an enabler of a more sustainable future when applied effectively \autocite{veitDigitalizationProblemSolution2023}. This duality underlines the importance of a clear and multi-dimensional understanding of \textit{sustainability}.

To contextualize the broader implications of AI deployment, it is essential to frame the discussion within the established model of sustainability, which provides a structured lens to assess environmental, economic, and social trade-offs. The widely accepted triad model of sustainability comprises three interdependent pillars \autocite{purvisThreePillarsSustainability2019}: (1) \textit{environmental} sustainability refers to preserving ecosystems and reducing emissions; (2) \textit{economic} sustainability involves long-term viability and cost-efficiency of systems; and (3) \textit{social} sustainability encompasses equitable access to resources, autonomy, and data control \autocite{klessascheckSOPAFrameworkSustainabilityoriented2025}. The increasing adoption of AI, particularly LLMs, introduces sustainability challenges across all three pillars \autocite{bossertWhyCarbonFootprint2025}. On the environmental side, high energy demands and carbon emissions during training and inference contribute to AI's growing ecological footprint \autocite{schutzeProblemSustainableAI2024, cowlsAIGambitLeveraging2023, liMakingAILess2023}. Economically, proprietary cloud-hosted models incur substantial operational costs, limiting scalability and excluding smaller organizations \autocite{fioravanteBusinessCaseResponsible2024, hoganAIHotMess2024}. Socially, centralized AI infrastructures raise concerns over digital sovereignty, data ownership, and fairness in access to powerful tools \autocite{lazarCanLLMsAdvance2024, sarkerDemocratizingKnowledgeCreation2024}.

Beyond technical performance, the choice between centralized and decentralized LLM deployment models carries significant implications for data governance, transparency, and organizational autonomy. While centralized models offer state-of-the-art performance, they may reinforce power asymmetries by consolidating control over infrastructure and data flows. In contrast, smaller or open-weight models—especially when deployed locally—can foster transparency, autonomy, and control over data. Moreover, from a data governance perspective, locally hosted LLMs can become essential in scenarios where sensitive information must remain under strict organizational control. Avoiding cloud-based inference pipelines helps minimize risks associated with data leakage, external surveillance, or jurisdictional conflicts related to data residency and compliance.

Despite increasing awareness of these multifaceted concerns, practical guidance for selecting sustainable AI models in organizational settings remains underdeveloped \autocite{kotlarskyDigitalSustainabilityInformation2023}. In the following, we translate this conceptual foundation into practical tooling decisions, comparing common LLM deployment options and their implications for sustainable use.

\subsection{Sustainability Considerations in LLM Tooling}

The trend toward deploying increasingly large, proprietary LLMs hosted on cloud platforms (e.g., GPT-4o and Claude-3 models) has intensified sustainability concerns due to their substantial energy demands, high operational costs, and infrastructure centralization \autocite{leonEscalatingAIsEnergy2024, singhSurveySustainabilityLarge2025}. These models typically incur high per-token costs, as illustrated in Table~\ref{tab:costs}, creating significant financial and environmental burdens for organizations seeking broad deployment. Conversely, smaller open-weight models such as Llama-3.3 or Phi-4, which can operate locally on existing organizational hardware, offer substantial economic and environmental advantages, dramatically reducing energy consumption per inference \autocite{wuSustainableAIEnvironmental2022, chenSurveyOnAI2023}. These models also enable greater flexibility for optimizing deployment, e.g., via quantization or model pruning, which further reduces compute cycles and energy use \autocite{verdecchiaASystematicReview2023}. Although these strategies incur additional emissions during training, operational savings often outweigh initial costs, especially in contexts that restrict cloud usage due to privacy or legal constraints \autocite{chenSurveyOnAI2023}.

From a cost standpoint, token pricing demonstrates stark disparities. As shown in Table~\ref{tab:costs}, input and output costs for proprietary models such as GPT-4.5-preview or Claude-3 can range from \$10 to over \$150 per million tokens. In contrast, smaller open-weight models like Llama-3.3-70B or Qwen-2.5-Coder are typically priced below \$1 per million tokens; we report the cost for Microsoft Azure deployment of these models per \url{https://models.litellm.ai/}. Notably, these figures represent hosted deployments; actual costs may be lower when models run locally on in-house infrastructure, particularly in organizations with idle compute resources. The difference in cost structures is not merely economic—it also implies environmental benefits through reduced compute intensity, aligning with green IT strategies \autocite{unhelkarEnterpriseGreenIT2012}.

\begin{table}[htbp]
\centering
\begin{tabular}{lrr}
\toprule
\textbf{Model} & \textbf{Input Cost (\$)} & \textbf{Output Cost (\$)} \\
\midrule
\multicolumn{3}{c}{\textbf{Proprietary Models}} \\
\midrule
GPT-4.5-preview & 75.00 & 150.00 \\
GPT-4o & 2.50 & 10.00 \\
Claude-3.7-Sonnet & 3.00 & 15.00 \\
Claude-3.5-Sonnet & 3.00 & 15.00 \\
Grok-2-Latest & 2.00 & 10.00 \\
Grok-Beta & 5.00 & 15.00 \\
o3-mini & 1.10 & 4.40 \\
Gemini-2.0-Flash & 0.10 & 0.40 \\
\midrule
\multicolumn{3}{c}{\textbf{Open-Weight Models}} \\
\midrule
DeepSeek-R1-Distill-Llama-70B & 0.75 & 0.99 \\
Llama-3.3-70B-Instruct & 0.72 & 0.72 \\
Qwen-2.5-Coder-32B & 0.18 & 0.18 \\
Mistral-Nemo-Instruct-2407 & 0.15 & 0.15 \\
Phi-4 & 0.13 & 0.50 \\
Gemma-3$^*$ & 0.13 & 0.50 \\
\bottomrule
\end{tabular} \\
\caption{\label{tab:costs}Comparison of Token Costs Across AI Models (per million tokens)}
\smallskip
\textbf{Note:} Prices are from \url{https://models.litellm.ai/} and \url{https://llm-stats.com/}. $^*$No data for Gemma-3 was available at the time of writing, we report Phi-4 pricing as a proxy due to similar size and deployment costs.
\end{table}

Beyond cost and energy efficiency, deploying open-weight models on-premises enables organizations to retain complete control over data flows, ownership, and residency. This is particularly relevant in sectors governed by strict compliance and privacy regulations (e.g., GDPR regulations, \cite{nayakGDPRCompliantChatGPT2024}). From a data sovereignty perspective, reliance on external APIs from global providers introduces risk, mainly when sensitive or regulated information is processed during inference. Locally hosted LLMs offer a viable pathway to mitigate these concerns while maintaining strategic autonomy over data infrastructure and usage policies.

These technical and economic factors may suggest that open-weight, locally hosted models are not only a feasible but often preferable option for sustainable LLM deployment in practice. They align with all three pillars of sustainability: lowering energy use, reducing cost, and enabling greater data control and compliance.

\subsection{Rethinking AI Benchmarking for Sustainable Deployment}

Current AI benchmarking frameworks predominantly assess models based on technical performance metrics such as accuracy, robustness, scalability, and generalization \autocite{liCrowdsourcedDataHighQuality2024, whiteLiveBenchChallengingContaminationFree2024}. While essential for academic and technical comparison, these evaluations often omit critical deployment factors such as energy consumption, operational cost, and data sovereignty—elements that play a decisive role in organizational adoption \autocite{bolon-canedoAReview2024}. As a result, traditional benchmarks offer limited guidance when selecting models for real-world use under sustainability constraints \autocite{tripathiEthicalPracticesArtificial2025}.

Emerging task-oriented benchmarks aim to address some of these gaps by evaluating reasoning capabilities and linguistic robustness. For example, SimpleBench assesses compositional skills in zero-shot settings \autocite{vaccaroWhenCombinationsHumans2024}. However, such approaches still focus on cognitive performance and do not account for broader factors like environmental impact, infrastructure costs, or regulatory context, which are crucial for organizational use.

Platforms such as LMArena.ai\footnote{\url{https://lmarena.ai/} [Accessed: 08/04/2024]}, developed by UC Berkeley's SkyLab, introduce a crowdsourced evaluation methodology using anonymous, randomized model comparisons and the Elo rating system to capture relative performance \autocite{chiangChatbotArenaOpen2024}. The approach to reflect user preferences across diverse tasks such as coding or instruction-following is a common way to define ``good enough'' of LLM output: the fit between LLM output and human output for a comparable task (e.g., \cite{virkEnhancingTrustLLMGenerated2024}) with the goal to mimic human performance, which is not perfect nor does it need to be perfect to be useful \autocite{ferreiraGoodEnoughRepresentationsLanguage2002}. Still, this approach suffers from two limitations: First, it lacks alignment with specific professional task contexts; second, it risks skewed results due to its participants' demographic or experiential bias.

In contrast, benchmarking efforts focused on model efficiency and deployment feasibility are beginning to emphasize alternative evaluation criteria. Techniques like post-training quantization, which reduces the precision of model weights, can drastically lower memory and compute requirements while preserving output quality to a large extent \autocite{yaoExploringPosttrainingQuantization2024}. For instance, recent assessments of quantized Llama-3.1 models show full accuracy recovery across academic and applied tasks, demonstrating their practical suitability \autocite{kurticWeRanHalf2024}. Similarly, structured evaluation of inference efficiency, energy use, and compliance-readiness is critical to understanding a model's sustainability profile.

These developments raise the central question: When is a smaller LLM sufficient to support occupational tasks effectively? While larger models tend to improve readability and informativeness, they may still introduce fidelity issues or generate hallucinated content \autocite{qiQuantifyingGeneralizationComplexity2025}. Smaller models may sometimes be more robust in high-constraint environments, e.g., when output must remain tightly faithful to input sources. This highlights the importance of task specificity: The optimal model depends not only on raw performance but also on how its capabilities align with task complexity and operational needs.

Neural scaling research further complicates the picture: larger models do not always yield better results unless trained with proportionally larger data sets and compute budgets. For example, the Chinchilla model (70B parameters) outperformed the larger Gopher model (280B) by optimizing the balance between model size and training dataset size \autocite{kaplanScalingLawsNeural2020, hoffmannTrainingComputeOptimalLarge2022a}. These findings caution against an overreliance on size alone as a proxy for capability and suggest a need for more nuanced evaluation standards; in fact, the recent Gemma-3 model family from Google achieved impressive performance, often rivaling much larger models \autocite{teamGemma3Technical2025}.

\section{Methodology}
\label{sec:method}

To systematically compare the performance of different LLMs across realistic occupational tasks, we developed an evaluation framework based on two LLMs as we detail below; cf.~Figure~1. This setup allows for scalable and standardized testing of multiple models on fixed prompts, with consistent evaluation using high-performing LLM. The basic idea behind this approach is that the high-performing LLMs will find output similar to their own answers and knowledge bases more acceptable, akin to the apprenticeship learning paradigm, see e.g., \autocite{grover2024better}. Using this approach, we evaluated a wide range of different LLMs and tested which ones deliver acceptable output quality (compared to their larger, more complex siblings) for typical organizational tasks while considering sustainability and cost dimensions.

\subsection{LLM-based Evaluation Setup}

Our system follows a two-component LLM-based testing and evaluation architecture: an \textit{execution component} generates output for a given task using a specific LLM, and an \textit{evaluation component} scores this output for 10 defined criteria using a fixed evaluation prompt.

Each task LLM is a model instance (e.g., GPT-4o, Llama-3.3, DeepSeek-R1) that receives a structured task prompt (cf.~Section~\nameref{sec:tasks}) and produces a one-shot output without post-processing or iterative refinement. 
The generated output is then rated by three evaluator LLMs (Gemini-2.0-Flash, GPT-4o, o3-mini), across ten predefined criteria (cf.~Section~\nameref{sec:eval}). Ratings were on a scale from 1 (poor) to 10 (excellent). This approach allowed for fully automated comparisons across all model-task combinations.

\vspace{-0.2cm}
\begin{figure}[h]
\centering
\includegraphics[width=0.87\textwidth]{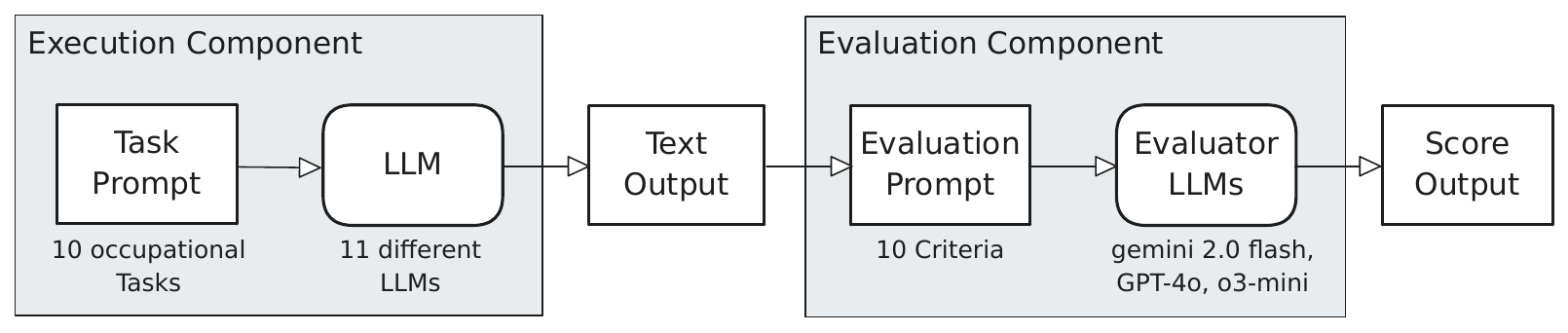}
\vspace{0.3em}

\textbf{\usefont{T1}{ptm}{b}{n}Figure 1.\hspace{0.09cm} Overview of the applied Testing and Evaluation Framework}
\caption*{\label{fig:framework}} 
\end{figure}
\vspace{-0.9cm}

\subsection{Tested Models}
\label{sec:models}

We evaluated 11 different LLMs, including both proprietary cloud-based models (e.g., GPT-4o, Claude-3.7, and Gemini-2.0-Flash) and open-weight models that can be run locally on a single A100 GPU (e.g., Llama-3.3, DeepSeek-R1, Gemma-3), effectively allowing for local deployment. Table~\ref{tab:LLMs} provides an overview of the models, including size, openness, and model access.

\begin{table}[h]
\centering
\resizebox{\textwidth}{!}{%
\begin{tabular}{lccclll}
\toprule
\textbf{Model Name} & \textbf{Size} & \textbf{\shortstack[m]{Open\\Weights}} & \textbf{Thinking} & \textbf{Provider} & \textbf{Access} \\
\midrule
GPT-4o & Large & \redcross & \redcross & OpenAI & OpenAI API \\
o3-mini & Small & \redcross & \greencheck & OpenAI & OpenAI API \\
\addlinespace
Claude-3.5-sonnet & Large & \redcross & \redcross & Anthropic & Anthropic API \\
Claude-3.7-sonnet & Large & \redcross & \redcross & Anthropic & Anthropic API \\
\addlinespace
Gemini-2.0-Flash & Medium & \redcross & \redcross & Google & Google API \\
Gemma-3-27b & 27B & \greencheck & \redcross & Google & google/gemma-3-27b-it \\
\addlinespace
DeepSeek-R1-70b & 70B & \greencheck & \greencheck & DeepSeek & deepseek-ai/DeepSeek-R1-Distill-Llama-70B\\
Llama-3.3-70B-Instruct & 70B & \greencheck & \redcross & Meta & meta-llama/Llama-3.3-70B-Instruct \\
Phi-4 & 14B & \greencheck & \redcross & Microsoft & microsoft/phi-4 \\
Qwen-2.5-7B-Instruct-1M & 7B & \greencheck & \redcross & Qwen & Qwen/Qwen2.5-7B-Instruct-1M \\
Mistral-Nemo-Instruct-2407 & 12B & \greencheck & \redcross & Mistral & mistralai/Mistral-Nemo-Instruct-2407 \\
\bottomrule
\end{tabular}
}
\caption{\label{tab:LLMs}Large Language Models used in the Experiments. For open-weight models, the \emph{Hugging Face Hub} identifier has been provided.}
\end{table}

\subsection{Occupational Tasks as Evaluation Prompts}
\label{sec:tasks}

To simulate realistic workplace scenarios, we created a set of ten occupational tasks commonly associated with AI assistance in knowledge work. We took inspiration for these tasks from surveys of how LLMs are widely used at the workplace \parencite{chkirbeneLargeLanguageModels2024, haaseAugmentingCoachingGenAI2025, gallagherDoesUsingMultiple2021}. 
The ten tasks can be found in Table~\ref{tab:task-overview}. These tasks were selected to span different types of cognitive demand, including \emph{aggregation} (e.g., meeting minutes), \emph{conceptual} generation (e.g., outlining a proposal), and \emph{transformation} (e.g., rewriting information). The logic is that any input is either condensed, enhanced, or transformed into a different form or output style through the LLM. Each task contained structured input material and a clear instruction format. The complete list of tasks, along with the prompting, is available in Appendix A on OSF\footnote{\url{https://osf.io/p5t8m/?view_only=c694110912c94936809d040def2555e8}}.

\begin{table}[h]
\centering
\small
\begin{tabular}{p{4.5cm}p{11.1cm}}
\toprule
\textbf{Task Title} & \textbf{Brief Description} \\
\midrule
\multicolumn{2}{l}{\textit{Aggregation Tasks}} \\
\midrule
Summarizing a Complex Text & Produce three levels of summary (sentence, executive summary, bullets) from a dense editorial policy text, balancing clarity and information preservation. \\
Generating Meeting Minutes & Transforming jumbled, informal notes into professional meeting minutes, including participants, agenda, discussion, and action items. \\
Extracting Key Information & Extract action-relevant information (goals, deadlines, open questions) from a long email and organizing it in a clear, scannable format. \\
Creating a Comparison Table & Building a comparison table (features, pros/cons, pricing, use cases) based on unstructured product info, followed by a short selection recommendation. \\
\midrule
\multicolumn{2}{l}{\textit{Conceptual Tasks}} \\
\midrule
Creating a Time Schedule & Generating a structured daily schedule based on task demands, energy levels, and meeting constraints, with justifications for time block choices. \\
Drafting a Project Proposal Outline & Turning a short campaign description into a structured proposal outline (title, goals, methods, timeline, budget) using a bullet-point format. \\
Writing a To-Do List Based on a Goal & Break down a vague event planning goal into actionable steps, grouped by phase, with suggested deadlines and explanatory notes. \\
Planning and Structuring a Presentation & Creating a compelling presentation outline (title, slide structure, opening hook) for a general audience, balancing accessibility and insight. \\
\midrule
\multicolumn{2}{l}{\textit{Transformation Tasks}} \\
\midrule
Writing an Email (Formal and Informal) & Composing two formal and informal emails based on the same situation, adjusting tone, style, and structure while preserving content. \\
Rewriting for Clarity and Tone & Rewrite a convoluted paragraph into three versions with different tones: neutral/professional, friendly/conversational, and concise/direct. \\
\bottomrule
\end{tabular}
\caption{Occupational Tasks used for Model Evaluation, Grouped by Task Category}
\label{tab:task-overview}
\end{table}

\subsection{Evaluation Criteria and Scoring}
\label{sec:eval}

To assess the quality of outputs generated by each LLM, we implemented a standardized evaluation procedure using three different LLMs as evaluators: \textit{GPT-4o}, \textit{Gemini-2.0-Flash}, and \textit{o3-mini}. This triad was selected to ensure robustness across evaluator types: GPT-4o represents a state-of-the-art, high-performance model, Gemini-2.0-Flash offers a lightweight, general-purpose option, and o3-mini exemplifies a compact ``thinking model'' architecture. Using multiple evaluators allows us to triangulate scoring consistency and assess potential biases introduced by individual LLM preferences.

Each evaluator received identical scoring prompts and independently rated the outputs of the eleven tested models across ten evaluation criteria (cf.~Table~\ref{tab:evaluations}). These criteria builds on prior work in LLM evaluation \autocite{chiangChatbotArenaOpen2024, jiBeaverTailsImprovedSafety2023, longLLMsDrivenSyntheticData2024} and are adapted to reflect core dimensions of output quality especially relevant for occupational tasks: (1) \emph{core qualities}, (2) \emph{factual integrity}, and (3) \emph{ethical and social responsibility}.

\begin{table}[hb!]
\centering
\small
\begin{tabular}{p{4.1cm}p{3.3cm}p{7.8cm}}
\toprule
\textbf{Category} & \textbf{Criterion} & \textbf{Explanation} \\
\midrule
\multirow{5}{*}{Core Qualities} 
 & Usefulness & Practical value of the response for the end-user \\
 & Relevance & Appropriateness of content given the question and inputs \\
 & Completeness & Extent to which the answer covers all necessary aspects \\
 & Clarity & Coherence and readability of the output \\
 & Tone Appropriateness & Appropriateness of tone for the target audience \\
\midrule
\multirow{2}{*}{Factual Integrity}
 & Accuracy & Correctness of factual content \\
 & Hallucination Freeness & Lack of invented or misleading content \\
\midrule
\multirow{3}{*}{Ethical \& Social Responsibility}
 & Ethical Bias & Presence of inclusive, non-discriminatory content \\
 & Political Bias & Absence of politically slanted or agenda-driven framing \\
 & Censorship & Avoidance of overly filtered or withheld responses \\
\bottomrule
\end{tabular}
\caption{Evaluation Criteria Grouped by Category}
\label{tab:evaluations}
\end{table}

Each criterion was scored on a 10-point Likert scale, where 1 indicated poor performance and 10 represented excellent performance. The evaluators processed outputs without knowledge of which model generated them and, importantly, also assessed their own outputs in a blind fashion. This introduces a potential for self-favoring bias. However, our comparative evaluation across the three evaluators (cf.~Figure~1 in Appendix B on OSF) shows that GPT-4o was not rated disproportionately higher by itself compared to Gemini-2.0-Flash or o3-mini. This is important as for the following inference statistics and analyses, we use GPT-4o only: using the intra-class correlation coefficient (ICC), a two-way random-effects model with absolute agreement was applied (ICC[2,3]). The resulting ICC was $<.50$, indicating bad agreement according to the interpretation guidelines by \textcite{kooGuidelineSelectingReporting2016}. Thus, we proceeded to report the main results using scores from GPT-4o as the primary evaluator with the most robust and reliable scores. This is in line with other recent work, where GPT-4o has been proposed as a gold standard for LLM-based evaluation due to its high alignment with human judgment \parencite{chiangChatbotArenaOpen2024} and comparably bad evaluation performance by other LLMs like Gemini-2.0-Flash and o3-mini \autocite{petrovProofBluffEvaluating2025}. Further, both Gemini-2.0-Flash and o3-mini show weaker performances across the tasks and overall very high positive evaluations across tasks (cf.~Figure~1 in Appendix B on OSF), lowering their potential to provide proper evaluations.  

Finally, each model-task combination was evaluated only once to reflect typical ``one-shot'' usage patterns in real-world applications. We intentionally avoided prompt engineering or multi-shot refinement to simulate how an average user or employee might interact with LLMs in everyday organizational settings. This ecological approach allows us to assess sufficiency rather than maximal performance and better informs the realistic applicability of each model across tasks.

\subsection{Analytical Strategy}
\label{sec:analytical-strategy}

We aggregated output scores across tasks to systematically compare the evaluated LLMs and conducted analyses at both the model and task levels. We worked with the plain scores from the 1-10 evaluations for the following analysis. For a more precise comparison between models across tasks, we applied $z$-transformation to all task scores because individual tasks varied in difficulty and output expectations. This allowed us to compare relative model performance across tasks on a standardized scale and to interpret performance differences in terms of deviation from average task difficulty (cf.~Appendix C on OSF).

We pursued three complementary analytical steps to address our research questions:

\begin{enumerate}
\item \textbf{Model-level comparison:} We conducted one-way ANOVAs to test whether different models produced significantly different outputs in terms of overall quality. This analysis established whether model choice systematically influences the usability and quality of generated content, particularly in light of sustainability trade-offs (e.g., smaller vs. larger models).

\item \textbf{Task-type comparison:} To explore how performance varies by cognitive demand, we grouped tasks into three categories—\textit{aggregation}, \textit{transformation}, and \textit{conceptual}—and performed ANOVAs with post-hoc Games-Howell tests. This helped us identify which task types posed particular challenges for LLMs, and whether model suitability is task-specific.

\item \textbf{Cluster analysis:} We performed K-means clustering based on the ten evaluation criteria to group models into distinct performance profiles. This allowed us to go beyond individual criteria and detect overarching patterns—e.g., premium all-rounders vs. reliable minimalists—supporting decision-making for context-sensitive model selection.
\end{enumerate}

All analyses reported in the main paper were based on evaluation scores generated by GPT-4o, which we used as a consistent, high-performance evaluator to visualize the evaluation pattern of all three evaluators. Supplementary analyses using the other evaluators are available in Appendix B on OSF. 

By combining standardized scoring, inferential statistics, and unsupervised pattern detection, this analytical strategy provides both detailed and holistic insights into LLM performance, informing the trade-offs between model size, sustainability, and sufficiency for occupational work.

\section{Results}
\label{sec:results}
We present the results of our comparative evaluation of eleven LLMs across ten everyday occupational tasks. Analyses were conducted at the level of evaluation dimensions, task categories, and individual tasks, and were complemented by a cluster analysis to identify model performance profiles.

\subsection{Performance Analysis across Evaluation Criteria}
Individual ANOVA for evaluation criteria across models revealed significant differences in the task performance, indicating that model choice significantly affects the quality of outputs across everyday workplace tasks (cf.~Table~\ref{tab:result_eval}). Importantly, despite these differences, average scores across models were generally high, reflecting the maturity of current LLMs for occupational applications. Breaking down performance by evaluation criterion based on GPT-4o ratings, we found statistically significant differences for: \textit{Accuracy}, $F(10, 99) = 2.79$, $p = .004$; \textit{Clarity}, $F(10, 99) = 2.47$, $p = .011$; \textit{Completeness}, $F(10, 99) = 2.21$, $p = .023$; \textit{Relevance}, $F(10, 99) = 3.22$, $p = .001$; \textit{Tone Appropriateness}, $F(10, 99) = 2.65$, $p = .007$; and notably, \textit{Usefulness}, $F(10, 99) = 3.48$, $p < .001$. These criteria reflect practical dimensions of output quality, especially usefulness, which captures whether human users can directly apply, adapt, or build upon the generated text. In this sense, usefulness acts as a proxy for ``operational sufficiency'' and may be especially relevant when considering task automation or AI augmentation in real-world settings.

In contrast, no significant differences emerged in criteria related to factual safety or ethics: \textit{Hallucination Freeness}, $F(10, 99) = 1.71$, $p = .089$; \textit{Ethical Bias}, $F(10, 99) = 0.08$, $p = 1.000$; \textit{Political Bias}, $F(10, 99) = 0.08$, $p = 1.000$; and \textit{Censorship}, $F(10, 99) = 0.10$, $p = 1.000$. This suggests that most models—even smaller open-weight ones—have achieved a robust baseline in minimizing inappropriate or misleading content, likely due to consistent alignment training and safety reinforcement across providers.

These results show that while LLMs are broadly capable across core dimensions, meaningful differences remain in how well they deliver user-relevant and context-sensitive outputs. These differences are particularly pronounced in criteria that directly impact downstream usability—underscoring the importance of task-aware evaluation beyond simple correctness or style.

\begin{table}[htbp]
\centering
\small
\begin{adjustbox}{width=\textwidth}
\begin{tabular}{llccccccccccc}
\toprule
 & \textbf{Model} 
 & \multicolumn{5}{c}{\textbf{Core Qualities}} 
 & \multicolumn{2}{c}{\textbf{Factual Integrity}} 
 & \multicolumn{3}{c}{\textbf{Responsibility}} \\
\cmidrule(lr){3-7} \cmidrule(lr){8-9} \cmidrule(lr){10-12}
& & Usefulness & Relevance & Completeness & Clarity & Tone & Accuracy & Halluc. & Eth. Bias & Pol. Bias & Censor. \\
\midrule
\multicolumn{12}{c}{\textbf{Proprietary Models}} \\
\midrule
& Claude-3.5-sonnet-latest & \cellcolor{orange2}8.70 & \cellcolor{yellow1}9.20 & \cellcolor{orange2}8.85 & \cellcolor{orange2}8.85 & \cellcolor{yellow1}9.35 & \cellcolor{yellow1}9.20 & \cellcolor{green2}9.60 & \cellcolor{green2}9.75 & \cellcolor{green2}9.80 & \cellcolor{green2}9.85 \\
& Claude-3.7-sonnet-latest & \cellcolor{yellow2}8.85 & \cellcolor{yellow1}9.25 & \cellcolor{orange2}8.90 & \cellcolor{yellow2}9.10 & \cellcolor{yellow1}9.40 & \cellcolor{yellow1}9.25 & \cellcolor{green2}9.75 & \cellcolor{green2}9.75 & \cellcolor{green2}9.85 & \cellcolor{green2}9.90 \\
& Gemini-2.0-Flash & \cellcolor{yellow2}8.85 & \cellcolor{yellow2}8.95 & \cellcolor{orange2}8.85 & \cellcolor{orange2}8.85 & \cellcolor{yellow2}9.15 & \cellcolor{orange2}9.05 & \cellcolor{green2}9.70 & \cellcolor{green2}9.80 & \cellcolor{green2}9.85 & \cellcolor{green1}9.95 \\
& GPT-4o & \cellcolor{green1}9.50 & \cellcolor{green1}9.70 & \cellcolor{green1}9.25 & \cellcolor{green1}9.30 & \cellcolor{green1}9.70 & \cellcolor{green1}9.40 & \cellcolor{green1}9.95 & \cellcolor{green1}9.85 & \cellcolor{green2}9.85 & \cellcolor{green2}9.90 \\
& o3-mini & \cellcolor{yellow2}8.90 & \cellcolor{yellow1}9.25 & \cellcolor{orange2}8.90 & \cellcolor{orange2}8.65 & \cellcolor{orange2}9.05 & \cellcolor{orange2}9.00 & \cellcolor{green2}9.65 & \cellcolor{green2}9.80 & \cellcolor{green2}9.80 & \cellcolor{green2}9.90 \\
\midrule
\multicolumn{12}{c}{\textbf{Open-Weight Models Locally Deployed}} \\
\midrule
& DeepSeek-R1-70b & \cellcolor{orange1}8.05 & \cellcolor{orange1}8.50 & \cellcolor{orange1}8.30 & \cellcolor{orange1}8.55 & \cellcolor{yellow2}9.00 & \cellcolor{orange1}8.55 & \cellcolor{green2}9.40 & \cellcolor{green2}9.75 & \cellcolor{green2}9.80 & \cellcolor{green2}9.90 \\
& Gemma-3-27b & \cellcolor{orange2}8.60 & \cellcolor{orange2}8.90 & \cellcolor{orange2}8.60 & \cellcolor{orange2}8.75 & \cellcolor{orange2}9.05 & \cellcolor{yellow2}8.90 & \cellcolor{green2}9.45 & \cellcolor{green2}9.75 & \cellcolor{green2}9.85 & \cellcolor{green2}9.90 \\
& Llama-3.3-70B-Instruct & \cellcolor{orange2}8.50 & \cellcolor{orange2}8.80 & \cellcolor{orange2}8.55 & \cellcolor{orange2}8.50 & \cellcolor{orange2}8.80 & \cellcolor{orange2}8.70 & \cellcolor{yellow2}9.15 & \cellcolor{green2}9.70 & \cellcolor{green2}9.75 & \cellcolor{green2}9.85 \\
& Mistral-Nemo-Instruct-2407 & \cellcolor{orange1}8.15 & \cellcolor{orange1}8.50 & \cellcolor{orange2}8.40 & \cellcolor{orange1}8.35 & \cellcolor{orange2}8.80 & \cellcolor{orange2}8.55 & \cellcolor{green2}9.35 & \cellcolor{green2}9.75 & \cellcolor{green2}9.80 & \cellcolor{green2}9.85 \\
& Phi-4 & \cellcolor{orange2}8.65 & \cellcolor{orange2}8.85 & \cellcolor{orange2}8.70 & \cellcolor{orange2}8.75 & \cellcolor{orange2}9.05 & \cellcolor{yellow2}8.90 & \cellcolor{green2}9.40 & \cellcolor{green2}9.70 & \cellcolor{green2}9.80 & \cellcolor{green2}9.85 \\
& Qwen-2.5-7B-Instruct-1M & \cellcolor{orange1}8.35 & \cellcolor{orange2}8.60 & \cellcolor{orange1}8.25 & \cellcolor{orange1}8.55 & \cellcolor{orange2}8.85 & \cellcolor{orange2}8.65 & \cellcolor{green2}9.40 & \cellcolor{green2}9.70 & \cellcolor{green2}9.75 & \cellcolor{green2}9.80 \\
\bottomrule
\end{tabular}
\end{adjustbox}
\caption{\label{tab:llm_scores}Evaluation of LLMs Across Core Qualities, Integrity, and Responsibility}
\label{tab:result_eval}
\end{table}

\subsection{Task Category Effects}

We computed an overall evaluation score per model by averaging across the ten tasks (cf.~Table~\ref{tab:scores}). A one-way ANOVA confirmed significant performance differences, $F(10, 99) = 3.51$, $p < .001$, with an effect size of $\eta^2 = .26$. This indicates that model selection explains approximately 26\% of the variance in output quality.

To test whether LLM performance varied across different types of tasks, we grouped the ten tasks into three categories: \textit{Aggregation tasks} (e.g., summarizing, listing); \textit{Transformation tasks} (e.g., rewriting, reorganizing), and \textit{Conceptual tasks} (e.g., ideation, proposal drafting). 

A one-way ANOVA with task category as the independent variable and overall evaluation score as the dependent variable revealed a significant effect, $F(2, 107) = 15.80$, $p < .001$, with a large effect size ($\eta^2 = .228$). Levene's test indicated unequal variances ($p = .002$), so Games-Howell post-hoc tests were used. These revealed that: \textit{Conceptual} tasks received significantly lower scores than both \textit{Aggregation} ($p < .001$) and \textit{Transformation} tasks ($p = .012$). No significant difference was found between \textit{Aggregation} and \textit{Transformation} tasks ($p = .624$). These results show that LLMs struggled most with open-ended conceptual prompts, suggesting limits in ``generative reasoning'' or ``creative structuring''.

\begin{table}[htbp]
\centering
\small
\begin{tabular}{lcccc}
\toprule
\textbf{Model} & \textbf{Overall} & \textbf{Aggregation} & \textbf{Conceptual} & \textbf{Transformation} \\
\midrule
\multicolumn{5}{c}{\textbf{Proprietary Models}} \\
\midrule
Claude-3.5-sonnet       & \cellcolor{yellow2}9.16 & \cellcolor{yellow2}9.05 & \cellcolor{yellow2}9.13 & \cellcolor{yellow1}9.25 \\
Claude-3.7-sonnet-latest      & \cellcolor{yellow1}9.25 & \cellcolor{yellow1}9.15 & \cellcolor{yellow1}9.15 & \cellcolor{yellow1}9.40 \\
Gemini-2.0-Flash              & \cellcolor{yellow1}9.18 & \cellcolor{green2}9.40 & \cellcolor{orange2}8.83 & \cellcolor{green2}9.43 \\
GPT-4o                         & \cellcolor{green1}9.62 & \cellcolor{green1}9.55 & \cellcolor{green1}9.45 & \cellcolor{green1}9.83 \\
o3-mini                       & \cellcolor{yellow2}9.14 & \cellcolor{yellow1}9.15 & \cellcolor{orange2}8.80 & \cellcolor{green2}9.48 \\
\midrule
\multicolumn{5}{c}{\textbf{Open-Weight Models Locally Deployed}} \\
\midrule
DeepSeek-R1-70b         & \cellcolor{orange2}8.71 & \cellcolor{orange2}8.70 & \cellcolor{orange1}8.23 & \cellcolor{yellow1}9.20 \\
Gemma-3-27b              & \cellcolor{orange2}8.92 & \cellcolor{orange2}8.85 & \cellcolor{orange2}8.73 & \cellcolor{yellow2}9.15 \\
Llama-3.3-70B-Instruct  & \cellcolor{orange2}8.82 & \cellcolor{orange1}8.50 & \cellcolor{orange2}8.68 & \cellcolor{yellow2}9.13 \\
Mistral-Nemo-Instruct-2407 & \cellcolor{orange2}8.68 & \cellcolor{orange2}8.70 & \cellcolor{orange1}8.18 & \cellcolor{yellow2}9.18 \\
Phi-4                  & \cellcolor{yellow2}8.98 & \cellcolor{orange2}8.60 & \cellcolor{orange2}8.83 & \cellcolor{yellow1}9.33 \\
Qwen-2.5-7B-Instruct-1M & \cellcolor{orange2}8.75 & \cellcolor{orange1}8.35 & \cellcolor{orange2}8.50 & \cellcolor{yellow1}9.20 \\
\bottomrule
\end{tabular}
\caption{\label{tab:scores}Performance Scores Across Tasks and Task Categories (on 0–10 scale)}
\end{table}

\subsection{Task-Specific Performance Differences}

A one-way ANOVA was conducted to assess whether LLM performance varied significantly across specific tasks using \textit{overall\_score} as the dependent variable and \textit{task\_name} as the independent variable. The analysis revealed a significant main effect of task, $F(9, 100) = 8.14$, $p < .001$, indicating substantial differences in performance across tasks. The effect size was large, with $\eta^2 = .423$. For more detailed analyses, across all three evaluators, for each task and LLM, see Appendix B on OSF. 

Levene's test for homogeneity of variance was significant ($p < .001$), violating the assumption of equal variances; thus, the Games-Howell post-hoc test was used. The results showed that the task \textit{Summarizing a Complex Text} received significantly lower scores than several other tasks, including \textit{Writing an Email (Formal and Informal)}, \textit{Generating Meeting Minutes}, \textit{Rewriting Text for Clarity and Tone}, \textit{Extracting Key Information from a Document}, \textit{Creating a Comparison Table}, and \textit{Planning and Structuring a Presentation} ($p < .05$). Potentially because the text to be summarized required a relatively long input sequence. In contrast, the task \textit{Planning and Structuring a Presentation} received significantly higher scores than tasks such as \textit{Writing an Email}, \textit{Generating Meeting Minutes}, \textit{Rewriting Text}, \textit{Extracting Key Information}, and \textit{Creating a Comparison Table}. For a detailed comparison, based on z-scores for each LLM for all 10 tasks, see Appendix C on OSF.

These findings suggest that the specific task prompt had a notable impact on LLM output quality. Tasks requiring conceptualizations and idea generation posed more significant challenges for most models, and tasks like rewriting in style and tone appeared to better align with the capabilities of the models under evaluation.

\subsection*{Clustering of LLMs Based on Evaluation Profiles}

To identify overarching patterns in model performance, we conducted a K-means cluster analysis based on the ten evaluation criteria (\textit{Usefulness, Relevance, Completeness, Clarity, Tone Appropriateness, Accuracy, Hallucination Freeness, Ethical Bias, Political Bias,} and \textit{Censorship}). The analysis included the mean scores across all tasks for each LLM. A three-cluster solution was specified, which successfully grouped the eleven models into distinct performance profiles. The three-cluster solution grouped the models into clearly distinct sets: Cluster~1 (n = 1), Cluster~2 (n = 6), and Cluster~3 (n = 4). The K-means solution explained a large proportion of variance in key criteria, with high between-cluster variation in \textit{Usefulness, Completeness, Accuracy}, and \textit{Relevance}. A series of one-way ANOVAs was conducted to descriptively assess the differences in evaluation criteria across the three clusters. The analysis revealed large and statistically significant differences in most dimensions, including \textit{Usefulness}, $F(2, 8) = 28.01$, $p < .001$, completeness, $F(2, 8) = 24.56$, $p < .001$, \textit{Accuracy}, $F(2, 8) = 22.38$, $p < .001$, and \textit{Relevance}, $F(2, 8) = 19.44$, $p < .001$. More minor but still significant differences were observed for \textit{Hallucination Freeness, Clarity, Tone Appropriateness, Ethical Bias}, and \textit{Political Bias} ($p < .05$). No significant differences were found in \textit{Censorship} scores, which remained consistently high across all clusters.

To estimate the strength of these differences, we computed the proportion of variance explained by the clustering (expressed as $\eta^2$). The clustering explained 88\% of the variance in \textit{Usefulness} scores, 83\% in \textit{Completeness}, 82\% in \textit{Accuracy}, 71\% in \textit{Relevance}, and 68\% in \textit{Clarity}. These large effect sizes confirm that the clustering structure reflects divergent performance profiles across models. It is important to note that these ANOVAs serve descriptive purposes, as the clustering algorithm itself optimizes separation across groups. Nevertheless, they provide strong evidence that the identified clusters capture meaningful differences in LLM performance across evaluation dimensions.

\subsubsection{Cluster 1: Premium All-Rounder.}
This cluster contained a single model: \textit{GPT-4o}. It was characterized by the highest performance across all evaluation dimensions, including \textit{Clarity} ($M = 9.30$), \textit{Hallucination Freeness} ($M = 9.95$), usefulness ($M = 9.50$), and \textit{Relevance} ($M = 9.70$). GPT-4o showed no notable weaknesses, excelling in factual accuracy, stylistic dimensions, and ethical criteria. This model thus stands out as a premium, high-performance all-rounder.

\subsubsection{Cluster 2: Competent and Responsible Models.}
This group included \textit{Claude-3.5}, \textit{Claude-3.7}, \textit{Gemini-2.0-Flash}, \textit{Gemma-3}, \textit{o3-mini}, and \textit{Phi-4}. These models demonstrated solid overall performance, particularly in responsibility-related criteria such as \textit{Censorship} ($M = 9.89$), \textit{Ethical Bias}  ($M = 9.76$), and \textit{Tone Appropriateness} ($M = 9.17$). While slightly behind GPT-4o on \textit{Clarity}, \textit{Usefulness}, and \textit{Completeness}, they still maintained high levels of consistency across evaluation criteria. This cluster represents balanced, reliable models with responsible and stylistically sound output.

\subsubsection{Cluster 3: Limited but Safe Models.}
The third cluster consisted of \textit{DeepSeek}, \textit{Llama-3}, \textit{Mistral-Nemo-Instruct}, and \textit{Qwen-2.5}. These models received the lowest average scores across nearly all dimensions, particularly \textit{Usefulness} ($M = 8.26$), \textit{Clarity} ($M = 8.49$), and \textit{Completeness} ($M = 8.38$). While all these scores are relatively high, their output can in part be considered unuseful enough to require second runs in practice or a considerable effort by the human user to make it usable. Despite these limitations, they still performed relatively well on ethical criteria, including \textit{Censorship} ($M = 9.85$) and \textit{Political Bias} ($M = 9.78$). This group can be interpreted as containing technically and ethically safe models, yet underwhelming in terms of expressive output quality or task richness.

These findings provide further nuance to the comparative evaluation of LLMs, showing that while some models offer broad excellence across evaluation dimensions, others specialize in either responsibility or baseline task completion. The cluster profiles provide practical insights for selecting LLMs based on application priorities, such as creative fluency versus safety and bias mitigation.

\section{Discussion}
\label{sec:discussion}

In our context, ``good enough'' does not refer to reaching the absolute best possible output quality, but rather to whether an LLM produces results that are (a) functionally usable, (b) informationally sufficient, and (c) stylistically appropriate for professional downstream use. In other words, does the output enable a knowledge worker to continue or complete a task without major revisions, corrections, or risk exposure? This operational definition differs from generic benchmarking or human-substitution goals often found in crowd-based evaluations such as Chatbot Arena \parencite{chiangChatbotArenaOpen2024}, which measure relative preference across diverse tasks in anonymous settings. While such platforms offer useful performance gradients, they tend to conflate general helpfulness with task fitness and overlook domain constraints. Our approach therefore, grounds the evaluation in concrete occupational tasks with clear criteria, aligning ``good enough'' with practical utility in real-world workflows rather than abstract model superiority.

\subsection{Performance Versus Sufficiency: Task Matters, So Does Context}

Our results show that model choice does matter: significant performance differences were observed across the eleven tested LLMs. The model GPT-4o consistently outperformed all other models across core quality criteria (usefulness, clarity, completeness, accuracy), establishing itself as a strong all-rounder. Claude-3.7 also performed reliably, especially in tasks requiring structured planning or synthesis. These models represent the current benchmark for high-quality LLM support in business contexts. At the same time, the even more powerful ``thinking models'' like o3-mini or Claude-3.7-thinking are not that well suited for our test tasks, not because they are not effective but rather because they require a very different prompting approach, which is not akin to the dialog style most users are used to and employ; they are much more tailored to solving complex tasks with a single answer with very specific and extensive prompting\footnote{\url{https://www.latent.space/p/o1-skill-issue} [Accessed: 08/04/2025]}. At the same time, certain smaller (and open-weight) models, particularly Gemma-3 and Phi-4, achieved solid and stable performance. While they did not reach top-tier scores, they delivered consistent and usable results. This positions them as viable alternatives in settings where autonomy, cost-efficiency, or deployment control take priority over peak performance. In contrast, most of the open-weight models evaluated displayed consistently weak performance, rendering them less suitable for reliable occupational use. Specifically, Mistral-Nemo-Instruct, DeepSeek-R1, Llama-3.3, and Qwen-2.5 frequently received negative to strongly negative ratings across multiple tasks. Common deficiencies included difficulties in producing clear, structured outputs, inadequate information summarization, and poor clarity or tone in rewriting tasks.

It is also important to note that model performance varied not just by architecture but by task type. Conceptual prompts (e.g., ideation or synthesis) proved more challenging than structured or transformation-oriented tasks. This reinforces the need for task-aware model selection: some applications may be well-served by efficient smaller models, while others still demand the cognitive flexibility of more advanced systems. These patterns underline the necessity of careful selection of LLMs according to task requirements. For versatile everyday applications, GPT-4o and Claude-3.7 are highly recommended due to their consistent quality and adaptability. Gemini-2.0-Flash, Gemma-3, and Phi-4 offer niche advantages in certain use cases, but require cautious deployment to ensure alignment with their specific strengths. The remaining models, including o3-mini, Claude-3.5, Mistral-Nemo-Instruct, DeepSeek-R1, Llama-3.3, and Qwen-2.5, are generally not recommended for occupational tasks due to their frequent shortcomings in quality and reliability. Future research should explore targeted improvements for these weaker-performing models to enhance their task effectiveness.

\subsection{Reframing Sustainable AI Deployment: Cost, Carbon, and Control}

Our findings contribute to a broader rethinking of what sustainable AI deployment entails. While environmental sustainability is often framed in terms of carbon emissions, a more complete view across all three dimensions of sustainability also includes financial viability and data governance. These additional dimensions are especially salient when LLMs are deployed at scale in cost-sensitive, compliance-constrained, or infrastructure-limited environments.

Inference cost is a central, yet often overlooked, factor in operational sustainability. Unlike model training, which is a one-time event, inference occurs repeatedly—often millions of times—and accumulates substantial energy use and emissions over time \parencite{Luccioni2022BLOOM, Li2024SPROUT}. Yet estimating inference energy consumption is notoriously difficult. Key determinants include model size, GPU efficiency, batch size, hardware type, and system-level optimizations such as KV caching or quantization; see e.g., Aschenbrenner (\citeyear{aschenbrennerSituationalAwareness2024}). Empirical estimates suggest that inference for current LLMs consumes between 0.5 and 1.3 kWh per million tokens, resulting in 150-300 gCO\textsubscript{2}e under a typical energy mix \parencite{You2025ChatGPTenergy, Samsi2023Words2Watts}. Larger models like GPT-4 typically lie at the upper end of this range, while quantized and smaller models like Mistral-7B or LLaMA2-7B show lower per-token footprints. Importantly, these figures exclude the embodied emissions of hardware and vary depending on whether models are deployed in the cloud or on local infrastructure.

Cost and energy efficiency can also be optimized through deployment strategies. Hardware matters: H100 GPUs, for example, offer up to 4.6x the inference throughput of A100s \parencite{You2025ChatGPTenergy}. Similarly, quantization methods (e.g., INT8) and effective batching can reduce energy consumption significantly. However, poor utilization---as seen, e.g., in the case of one open deployment of BLOOM---can lead to high per-request emissions due to idle resource overhead \parencite{Luccioni2022BLOOM}. In practice, organizations must balance these cost and carbon considerations against deployment constraints. Public cloud APIs provide state-of-the-art performance but introduce privacy risks, data transfer costs, and potential legal barriers to inference on sensitive documents. Even hybrid architectures may exclude confidential datasets, as seen in AI assistants within public research institutions \parencite{weberFhGenieCustomConfidentialityPreserving2024}.

Based on the performance results from this study's analyses, especially the cluster analyses showed clear differences between proprietary models and open weights, with the former typically scoring better. Figure~2 visualizes the trade-off between overall performance and cumulative input/output cost across models, illustrating the performance differences of different LLMs. Gemini-2 has the best price/performance ratio, but with Google as the hosting company leveraging Google's in-house TPUs, which are optimized ASICs for ML loads. Gemma-3 and Phi-4, therefore, show the best combination of performance/price for the general user; in particular, both models are open-weight and can run locally on a single (mid-range to high-end) GPU. This underscores the value of locally deployable models that can run efficiently on in-house GPUs in terms of sustainability, data sovereignty, and cost-efficiency. Our findings show that such models offer sufficient performance while drastically reducing cost-per-token and improving organizational control over data flows.

\begin{figure}[h]
\vspace{-0.2cm}
\centering
\begin{adjustbox}{max width=\textwidth,center}
\begin{tikzpicture}
\begin{axis}[
    xlabel={Overall Score of LLM},
    ylabel={Sum of Input and Output Cost of LLM},
    xmin=8.6, xmax=9.8,
    ymin=0, ymax=24,
    width=14cm,
    height=9cm,
    grid=both,
    legend style={at={(0.5,-0.15)}, anchor=north, legend columns=3, draw=none}
]

\addplot[only marks, mark=star, mark size=5pt] coordinates {(9.62,12.50)};
\addlegendentry{Cluster 1: Premium\ \ \ }

\addplot[only marks, mark=square, mark size=4pt] coordinates {
  (9.16,18.00) (9.25,18.00) (9.18,0.50) (9.14,5.50) (8.98,0.63) (8.92,0.63)
};
\addlegendentry{Cluster 2: Responsible\ \ \ }

\addplot[only marks, mark=triangle, mark size=4pt] coordinates {
  (8.71,1.74) (8.82,1.44) (8.75,0.35) (8.68,0.30)
};
\addlegendentry{Cluster 3: Limited}

\node[font=\footnotesize, anchor=west, rotate=90, xshift=1mm] at (axis cs:9.62,12.50) {GPT-4o};
\node[font=\footnotesize, anchor=west, rotate=90, xshift=1mm] at (axis cs:9.16,18.00) {Claude-3.5};
\node[font=\footnotesize, anchor=west, rotate=90, xshift=1mm] at (axis cs:9.25,18.00) {Claude-3.7};
\node[font=\footnotesize, anchor=west, rotate=90, xshift=1mm] at (axis cs:9.18,0.50) {Gemini-2.0};
\node[font=\footnotesize, anchor=west, rotate=90, xshift=1mm] at (axis cs:9.14,5.50) {o3-mini};
\node[font=\footnotesize, anchor=west, rotate=90, xshift=1mm] at (axis cs:8.98,0.63) {Phi-4};
\node[font=\footnotesize, anchor=west, rotate=90, xshift=1mm] at (axis cs:8.92,0.63) {Gemma-3};
\node[font=\footnotesize, anchor=west, rotate=90, xshift=1mm] at (axis cs:8.71,1.74) {DeepSeek-R1};
\node[font=\footnotesize, anchor=west, rotate=90, xshift=1mm] at (axis cs:8.82,1.44) {Llama-3.3};
\node[font=\footnotesize, anchor=west, rotate=90, xshift=1mm] at (axis cs:8.75,0.35) {Qwen2.5};
\node[font=\footnotesize, anchor=west, rotate=90, xshift=1mm] at (axis cs:8.68,0.30) {Mistral-Nemo-Instruct};

\end{axis}
\end{tikzpicture}
\end{adjustbox}
\vspace{0.3em}

\textbf{Figure~2.\hspace{0.09cm} LLM Performance vs. Sum of Input + Output Cost (by Marker Shape)}
\vspace{-0.8cm}
\caption*{}
\label{fig:cost_performance}
\end{figure}

\subsection{Toward Task-Aware, Context-Driven Benchmarking}

Although our study does not yet offer a complete decision framework, it provides empirical grounding for a broader benchmarking perspective that aligns model choice with real-world goals and constraints. We argue that evaluating LLMs solely through technical benchmarks risks overlooking critical dimensions of sustainability. Instead, sufficiency must be evaluated contextually: what is ``good enough'' depends on the task, the infrastructure, the cost model, and the data environment.

Our LLM-based evaluation method—using fixed tasks, standardized criteria, and LLM-based scoring—offers a scalable approach to capture such trade-offs. By running each model-task pair once, we simulate real-world use rather than optimized performance. This design choice highlights the practical gaps between idealized capabilities and applied value. Ultimately, our findings support a shift toward task-aware, context-driven LLM selection. For certain creative or high-stakes applications, premium models like GPT-4o may be necessary. But for many tasks, especially those requiring privacy, control, and cost-efficiency, smaller models are already good enough. The future of sustainable AI will depend not just on improving model capabilities but on improving our ability to choose wisely.

\section{Limitations}

Several limitations should be noted when interpreting the results of this study. First, while our evaluation spanned eleven diverse LLMs and ten representative tasks, the experimental setup was intentionally limited to single-shot outputs without prompt optimization or few-shot prompting. This design choice enhances ecological validity--reflecting how many users interact with LLMs in practice--but may underrepresent the full performance potential of specific models under optimized conditions.

Second, all inference statistics like ANOVAs were conducted using GPT-4o as the sole evaluator. While GPT-4o has demonstrated strong capabilities in evaluation tasks (e.g. \autocite{chiangChatbotArenaOpen2024}), relying on a single LLM introduces potential biases based on its preferences or blind spots. We mitigated this by using clear, rubric-based scoring criteria. Still, future work should incorporate human raters or multiple evaluators to validate the robustness of LLM-based scoring.

Third, our study focused on output quality and sustainability indicators but did not include direct energy consumption or latency measurements. We relied on publicly available estimates for model efficiency and token-level costs. While useful for comparison, actual deployment costs and carbon footprints may vary significantly depending on infrastructure, batch size, and inference patterns.

Finally, our sufficiency argument remains exploratory. We do not claim that a single score threshold defines adequacy for all contexts. Instead, we highlight the need for contextualized evaluation strategies that take into account task complexity, user goals, and organizational constraints.

\section{Conclusion}

This paper contributes to the growing conversation on sustainable and context-aware deployment of LLM in organizational settings. A systematic comparison of eleven proprietary and open-weight LLMs across ten everyday tasks shows that smaller, locally deployable models can be ``good enough'' for many real-world use cases, particularly when data privacy, cost efficiency, and infrastructure autonomy are essential.

Our findings demonstrate that while models like GPT-4o offer best-in-class performance, specific open-weight models, such as Gemma-3 and Phi-4, can often provide consistent and sufficient output quality at a fraction of the cost and energy footprint. These insights support a more nuanced understanding of AI sufficiency that goes beyond technical benchmarking to include sustainability, sovereignty, and practicality.

We argue that evaluating LLMs solely through traditional metrics risks overlooking critical operational considerations. Our agent-based evaluation method offers a replicable and scalable approach to model assessment, and our results suggest the value of task-aware, context-driven model selection.

Future work should extend this approach by developing sufficiency frameworks incorporating task characteristics, infrastructure constraints, and user needs. In some cases, the most sustainable choice may not be to deploy an LLM at all---a decision that requires as much empirical grounding as adoption. As LLMs become embedded in organizational life, responsible deployment must begin with asking not only which model is best, but which model is \emph{right}.

\section{Acknowledgments}

The authors' work was partially supported by the German Federal Ministry of Education and Research (BMBF), grant number 16DII133 (Weizenbaum-Institute), by Deutsche Forschungsgemeinschaft under grants 496119880 (VisualMine), 531115272 (ProImpact), and SFB 1404/2 (FONDA), as well as Deutsche Forschungsgemeinschaft (DFG) through the DFG Cluster of Excellence MATH+ (grant number EXC-2046/1, project ID 390685689), as well as the \href{https://www.zib.de}{Zuse Institute Berlin} via the RISE@ZIB services hosting the LLM models.


\printbibliography

\end{document}